\documentclass{ieeeaccess}

\usepackage{cite}
\usepackage{amsmath,amssymb,amsfonts}
\usepackage{algorithmic}
\usepackage{graphicx}
\usepackage{textcomp}
\usepackage{bm}
\usepackage{booktabs}
\usepackage{tabularx}
\usepackage{float}
\usepackage{changepage}
\usepackage{changepage}
\newlength{\extralength}
\usepackage{lineno}
\usepackage{hyperref}

\makeatletter
\AtBeginDocument{\DeclareMathVersion{bold}
\SetSymbolFont{operators}{bold}{T1}{times}{b}{n}
\SetSymbolFont{NewLetters}{bold}{T1}{times}{b}{it}
\SetMathAlphabet{\mathrm}{bold}{T1}{times}{b}{n}
\SetMathAlphabet{\mathit}{bold}{T1}{times}{b}{it}
\SetMathAlphabet{\mathbf}{bold}{T1}{times}{b}{n}
\SetMathAlphabet{\mathtt}{bold}{OT1}{pcr}{b}{n}
\SetSymbolFont{symbols}{bold}{OMS}{cmsy}{b}{n}
\renewcommand\boldmath{\@nomath\boldmath\mathversion{bold}}}
\makeatother

\def\BibTeX{{\rm B\kern-.05em{\sc i\kern-.025em b}\kern-.08em
    T\kern-.1667em\lower.7ex\hbox{E}\kern-.125emX}}

\begin{document}

\history{}
\doi{}

\title{SWINSleepNet: A Hierarchical Context-Aware Framework for Sleep Staging (v2)}

\author{\uppercase{Chongjian Wang}\authorrefmark{1} and \uppercase{Junjie Gao}\authorrefmark{2}}

\address[1]{School of Mathematics and Systems Science, Shandong University of Science and Technology, Qingdao 266590, Shandong, China (e-mail: 202311080223@sdust.edu.cn)}
\address[2]{School of Artificial Intelligence, Shandong Women's University, Jinan 250352, Shandong, China (e-mail: junjie.gao@sdwu.edu.cn)}

\markboth
{Wang and Gao: SwinSleepNet: A Hierarchical Context-Aware Framework for Sleep Staging}
{Wang and Gao: SwinSleepNet: A Hierarchical Context-Aware Framework for Sleep Staging}

\corresp{Corresponding author: Junjie Gao (e-mail: junjie.gao@sdwu.edu.cn).}

\begin{abstract}
Automatic sleep staging is a critical role in sleep disorder diagnosis, sleep quality assessment, and long-term health monitoring; however, its performance remains limited in ambiguous and transition-related stages due to insufficient modeling of fine-grained intra-epoch structures and complex dependencies across different spectral regions. Most existing methods rely on conventional epoch-level encoders, which often fail to capture subtle temporal microstructures and cross-region interactions within a single epoch, leading to degraded performance on challenging stages such as N1. To address these limitations, we propose SwinSleepNet, a hierarchical context-aware dual-stream framework that explicitly decouples intra-epoch representation learning from inter-epoch contextual modeling. Specifically, each epoch is represented from two complementary views, namely the raw time-domain signal and its corresponding time–frequency representation. The raw branch employs a convolutional encoder to capture waveform-level temporal details, while the time–frequency branch leverages a Swin Transformer to model local spectro-temporal patterns, hierarchical multi-scale structures, and long-range dependencies across regions. The extracted features are fused into a unified embedding and further refined by a bidirectional contextual modeling module to incorporate temporal dependencies across epochs for final stage prediction. Extensive experiments on Sleep-EDF-20, Sleep-EDF-78, and SHHS demonstrate that the proposed framework achieves strong overall performance and yields more robust and stable recognition of challenging stages, particularly N1 and transition-related epochs, highlighting the importance of enhanced intra-epoch representation learning within hierarchical frameworks for automatic sleep staging.
\end{abstract}

\begin{keywords}
automatic sleep staging, EEG, time--frequency representation, Swin Transformer, dual-stream framework, hierarchical modeling, contextual learning
\end{keywords}

\titlepgskip=-15pt

\maketitle

\section{Introduction}

Sleep staging is a fundamental component of sleep medicine and sleep health assessment, serving as the basis for the diagnosis of sleep disorders, the evaluation of sleep quality, and long-term personalized health monitoring \cite{ref1,ref2}. Polysomnography (PSG), which simultaneously records multiple physiological signals such as electroencephalography (EEG), electrooculography (EOG), and electromyography (EMG), remains the clinical gold standard for sleep stage annotation \cite{ref2}. However, manual epoch-by-epoch scoring is labor-intensive, time-consuming, and highly dependent on expert experience, which limits its scalability in large clinical studies, home monitoring, and wearable health applications \cite{ref1,ref2}. These limitations have driven sustained interest in accurate, robust, and automated sleep staging systems \cite{ref1,ref7,ref10}.

Recent advances in deep learning have significantly improved automatic sleep staging \cite{ref13,ref16,ref20}. Convolutional neural networks (CNNs) have been widely used to learn discriminative features directly from raw EEG or time--frequency representations, thereby alleviating the dependence on handcrafted feature engineering \cite{ref3,ref13,ref16,ref18}. Subsequent studies further incorporated recurrent neural networks (RNNs), long short-term memory networks (LSTMs), and gated recurrent units (GRUs) to model temporal dependencies across consecutive epochs, improving sequence consistency in overnight sleep staging \cite{ref12,ref17,ref20}. More recently, Transformer-based architectures have been introduced to capture long-range contextual relationships through self-attention mechanisms \cite{ref19,ref20,ref21}. In parallel, multi-branch and multi-view frameworks have also emerged, aiming to exploit complementary information from raw waveforms, time--frequency representations, or multiple physiological modalities \cite{ref12,ref15,ref17}.

Despite this progress, several important challenges remain insufficiently addressed. First, automatic sleep staging is inherently a \emph{hierarchical} problem. On the one hand, the model must identify discriminative local structures within each epoch, such as sleep spindles, K-complexes, slow-wave activity, and rhythm variations. On the other hand, it must also leverage inter-epoch contextual information to characterize the temporal continuity and transition patterns of sleep stages \cite{ref12,ref20,ref21}. Although existing methods have increasingly recognized the importance of contextual modeling, epoch-level representation learning is still largely based on conventional convolutional or recurrent modules \cite{ref12,ref16,ref20}. While such modules are effective for local feature extraction, they remain limited in modeling multi-scale local structures and cross-region dependencies within a single epoch. Second, transitional stages such as N1 are short-lived, weakly discriminative, and highly overlapping with neighboring stages, making them particularly difficult to classify accurately \cite{ref5,ref16}. Shallow epoch encoders or single-scale feature extractors often fail to capture such subtle and ambiguous patterns \cite{ref16,ref17,ref19}. Third, in dual-stream or multi-view frameworks, the final performance critically depends on the quality of branch-specific representations and the effectiveness of feature fusion \cite{ref12,ref15,ref17}. If epoch-level representations are not sufficiently informative, the potential benefits of contextual modeling and multi-view complementarity cannot be fully realized.

From the perspective of model design, prior work has shown that decomposing sleep staging into \emph{intra-epoch representation learning} and \emph{inter-epoch contextual modeling} is both physiologically meaningful and empirically effective \cite{ref12,ref14,ref15}. In this paradigm, a model first extracts epoch-wise features and then refines them through sequential modeling across neighboring epochs. Such a hierarchical strategy is particularly appealing in dual-view settings, where raw time-domain signals and time--frequency representations provide complementary observations of the same underlying sleep state: the former preserves fine-grained waveform morphology, whereas the latter emphasizes spectral dynamics over time \cite{ref12,ref15,ref17}. Nevertheless, existing hierarchical frameworks still rely heavily on conventional epoch encoders, such as CNNs, BiLSTMs, or simple attention pooling modules, whose ability to capture local structures, multi-scale patterns, and cross-region dependencies within each epoch remains limited \cite{ref12,ref14,ref15}. Therefore, a key question is no longer merely whether contextual information should be modeled, but rather how to strengthen \emph{epoch-level representation learning} while preserving the benefits of hierarchical contextual modeling.

To address this issue, we propose a hierarchical context-aware dual-stream framework for automatic sleep staging. The proposed method preserves the two-level modeling paradigm of sleep staging, i.e., epoch-wise representation learning followed by inter-epoch contextual modeling, while introducing a Swin Transformer\cite{ref23} as the key epoch encoder to enhance the modeling of local structural patterns and cross-region dependencies within each epoch. Specifically, we employ a Swin Transformer-based hierarchical encoder in the time--frequency branch to extract more discriminative epoch-level representations with stronger local awareness and multi-scale modeling capability. These representations are then combined with complementary features from the raw-signal branch, followed by inter-epoch contextual modeling over consecutive sleep epochs. In this way, the proposed framework retains the advantages of hierarchical contextual modeling while substantially improving the quality of single-epoch representations, thereby supporting more reliable recognition of difficult sleep stages and ambiguous transition segments.

We evaluate the proposed method on multiple public sleep staging datasets under strict subject-wise protocols. Experimental results demonstrate that our framework achieves competitive overall performance in terms of accuracy, macro-F1, and Cohen's $\kappa$, while showing more stable recognition of difficult stages, especially N1 and transition-related epochs.These findings suggest that strengthening epoch-level representation learning within a hierarchical sleep staging framework is an effective direction for improving both discriminative performance and model robustness.

The main contributions of this work are summarized as follows:
\begin{itemize}
\item We propose a hierarchical context-aware dual-stream framework for automatic sleep staging, which explicitly decouples intra-epoch representation learning from inter-epoch contextual modeling and better reflects the hierarchical nature of the task.
\item We introduce a Swin Transformer as the key epoch encoder to enhance the modeling of local time--frequency structures, multi-scale patterns, and cross-region dependencies within individual sleep epochs.
\item We integrate dual-stream feature learning and sequential context modeling to exploit the complementarity between raw signals and time--frequency representations, improving overall sleep staging performance as well as the recognition of difficult stages such as N1.
\item We conduct systematic experiments on multiple public datasets, demonstrating the effectiveness and stability of the proposed framework across different sleep staging settings.
\end{itemize}

\section{Related Work}

\subsection{Deep Learning Frameworks for Automatic Sleep Staging}

Early automatic sleep staging systems mainly relied on handcrafted features extracted from time, frequency, and time--frequency domains, followed by shallow classifiers such as support vector machines and random forests \cite{ref8,ref9}. Although these methods provided a certain degree of interpretability, their performance was strongly constrained by the completeness and discriminability of manually designed features, and they were often sensitive to noise, inter-subject variability, and heterogeneous recording conditions.

The introduction of deep learning shifted the field toward end-to-end representation learning. CNN-based models such as DeepSleepNet, TinySleepNet, AttnSleep, DilatedSleepNet, and SleepEEGNet demonstrated that hierarchical convolutional filters can effectively capture discriminative temporal and spectral patterns directly from raw EEG or transformed inputs \cite{ref10,ref11,ref12,ref13,ref7}. These models significantly reduced the reliance on handcrafted features and established CNNs as a dominant backbone for sleep staging. In particular, multi-scale convolution, dilated convolution, and attention-enhanced convolutional designs improved sensitivity to characteristic sleep events such as sleep spindles, K-complexes, and slow-wave activity \cite{ref12,ref13}.

However, CNN-based frameworks are primarily optimized for local pattern extraction within individual epochs. Since sleep staging is not an isolated epoch classification problem but rather a sequential decision task governed by physiological transition regularities, purely convolutional models may not fully exploit dependencies across neighboring epochs. As a result, they often exhibit limited performance on difficult stages such as N1 and on ambiguous transition periods, where contextual information is especially important \cite{ref5,ref14}.

\subsection{Hierarchical and Context-Aware Sleep Staging Methods}

To better model temporal dependencies across sleep sequences, a large body of work has adopted hierarchical or context-aware frameworks that combine epoch-level feature extraction with sequence modeling. In these methods, a front-end encoder first produces an embedding for each epoch, and a back-end sequential module then models inter-epoch dependencies using LSTMs, GRUs, or related recurrent architectures. Representative examples include SeqSleepNet, XSleepNet, and MVF-SleepNet \cite{ref14,ref12,ref15}. Compared with pure CNN-based approaches, such frameworks explicitly incorporate sleep-stage transition dynamics and therefore tend to produce more temporally coherent predictions.

SeqSleepNet is a representative sequence-to-sequence framework that models sleep staging over a sequence of consecutive epochs rather than treating each epoch independently \cite{ref14}. XSleepNet further extends this idea by introducing a multi-view hierarchical framework, where raw time-domain signals and time--frequency representations are processed through separate branches before contextual modeling and final prediction \cite{ref12}. This line of work highlights an important methodological insight: automatic sleep staging should be treated as a \emph{two-level problem}, requiring both intra-epoch structural representation and inter-epoch contextual reasoning. Similarly, hybrid frameworks such as MVF-SleepNet also demonstrate that combining branch-specific feature extraction with sequence modeling can achieve a strong balance between overall accuracy and difficult-stage recognition \cite{ref15}.

Despite their success, most hierarchical frameworks still rely on conventional epoch encoders such as CNNs, BiLSTMs, or simple attention pooling modules. As a consequence, they are effective in handling \emph{inter-epoch contextual modeling}, but the quality of \emph{epoch-level representations} remains constrained by the expressive capacity of traditional front-end encoders. In other words, while these frameworks have established a strong contextual modeling paradigm, the problem of how to further improve single-epoch representation learning remains insufficiently explored \cite{ref12,ref14,ref15}.

\subsection{Transformer-Based Sleep Staging Models}

Transformer-based architectures have recently attracted increasing attention in automatic sleep staging due to their ability to model long-range dependencies through self-attention. Representative methods such as SleepTransformer, SleepViTransformer, and CNN--Transformer hybrids such as MixSleepNet and SalientSleepNet have shown that self-attention can effectively enhance sequence modeling and feature interaction in overnight sleep recordings \cite{ref19,ref20,ref18,ref21}. In addition, multi-channel Transformer variants have been explored to better capture contextual and inter-channel relationships in more complex PSG settings \cite{ref19,ref21}.

From a structural perspective, existing Transformer-based sleep staging methods can be broadly divided into two categories. The first category replaces or augments recurrent sequence modelers with self-attention modules, thereby strengthening the modeling of long-range inter-epoch context. The second category combines convolutional front-ends with Transformer layers, aiming to preserve local feature extraction while leveraging global context modeling. Overall, these methods represent an important step beyond CNN--RNN frameworks by integrating local structural cues with more flexible long-range dependency modeling \cite{ref18,ref19,ref20,ref21}.

Nevertheless, current Transformer-based approaches also exhibit limitations. Standard global self-attention often incurs considerable computational overhead, which may be suboptimal for medium-scale clinical datasets or resource-constrained settings. More importantly, many Transformer-based sleep staging methods focus primarily on \emph{sequence-level global modeling}, while paying less attention to \emph{how local structural patterns within individual epochs are encoded}. Since sleep staging decisions often depend on subtle and localized waveform or time--frequency events, a model that overemphasizes global sequence interaction without sufficiently strengthening epoch-level structural representation may still struggle with difficult stages and weakly discriminative transition segments \cite{ref19,ref20,ref21}.

\subsection{Multi-View and Multi-Branch Sleep Staging Methods}

Multi-view and multi-branch learning has become another important direction in automatic sleep staging. The central idea is to exploit complementary information from different representation spaces or physiological modalities, such as raw EEG waveforms, spectrogram-like time--frequency images, or multiple PSG channels. This strategy is motivated by the observation that different views emphasize different aspects of sleep physiology: raw signals preserve fine-grained waveform morphology, whereas time--frequency representations reveal the temporal evolution of spectral content. By learning from both views, the model may obtain a richer and more discriminative representation of each sleep epoch \cite{ref12,ref15,ref17}.

Existing multi-branch frameworks have demonstrated the potential value of such complementarity \cite{ref12,ref15,ref17}. However, multi-view learning does not automatically lead to better performance. In practice, the effectiveness of a dual-stream or multi-branch model depends critically on the quality of branch-specific representations, the degree of inductive diversity across branches, and the design of the fusion mechanism. If one branch fails to provide sufficiently informative epoch-level features, then the potential gains from multi-view interaction and subsequent contextual refinement remain limited. Therefore, the key challenge in multi-view sleep staging is not simply whether multiple branches are used, but whether each branch is equipped with a sufficiently strong epoch encoder and whether complementary information can be integrated effectively.

\subsection{Motivation of This Work}

Taken together, existing studies have established several important findings. First, contextual modeling across neighboring epochs is essential for sleep staging \cite{ref12,ref14,ref15}. Second, multi-view or multi-branch designs can provide complementary information and improve discriminative power \cite{ref12,ref15,ref17}. Third, self-attention-based architectures are promising for modeling complex dependencies in sleep data \cite{ref18,ref19,ref20,ref21}. However, a critical gap remains: while current hierarchical sleep staging frameworks have become increasingly effective at modeling \emph{inter-epoch context}, they still do not sufficiently strengthen \emph{intra-epoch representation learning}. In dual-stream settings, the quality of downstream contextual modeling and feature fusion is fundamentally bounded by the quality of the epoch-level embeddings produced by each branch.

The present work is motivated by this gap. Rather than discarding the hierarchical contextual modeling paradigm, we seek to strengthen its most crucial component, namely epoch-level representation learning. Considering the ability of the Swin Transformer to model local structures, cross-region interactions, and hierarchical multi-scale representations through window-based self-attention, we introduce it into a dual-stream sleep staging framework as a key epoch encoder. In this way, the proposed method aims to enhance single-epoch representations while preserving the benefits of contextual modeling and dual-view complementarity, ultimately yielding more accurate and stable automatic sleep staging.

\begin{figure*}[t]
\begin{adjustwidth}{-\extralength}{0cm}
\centering
\includegraphics[width=\textwidth]{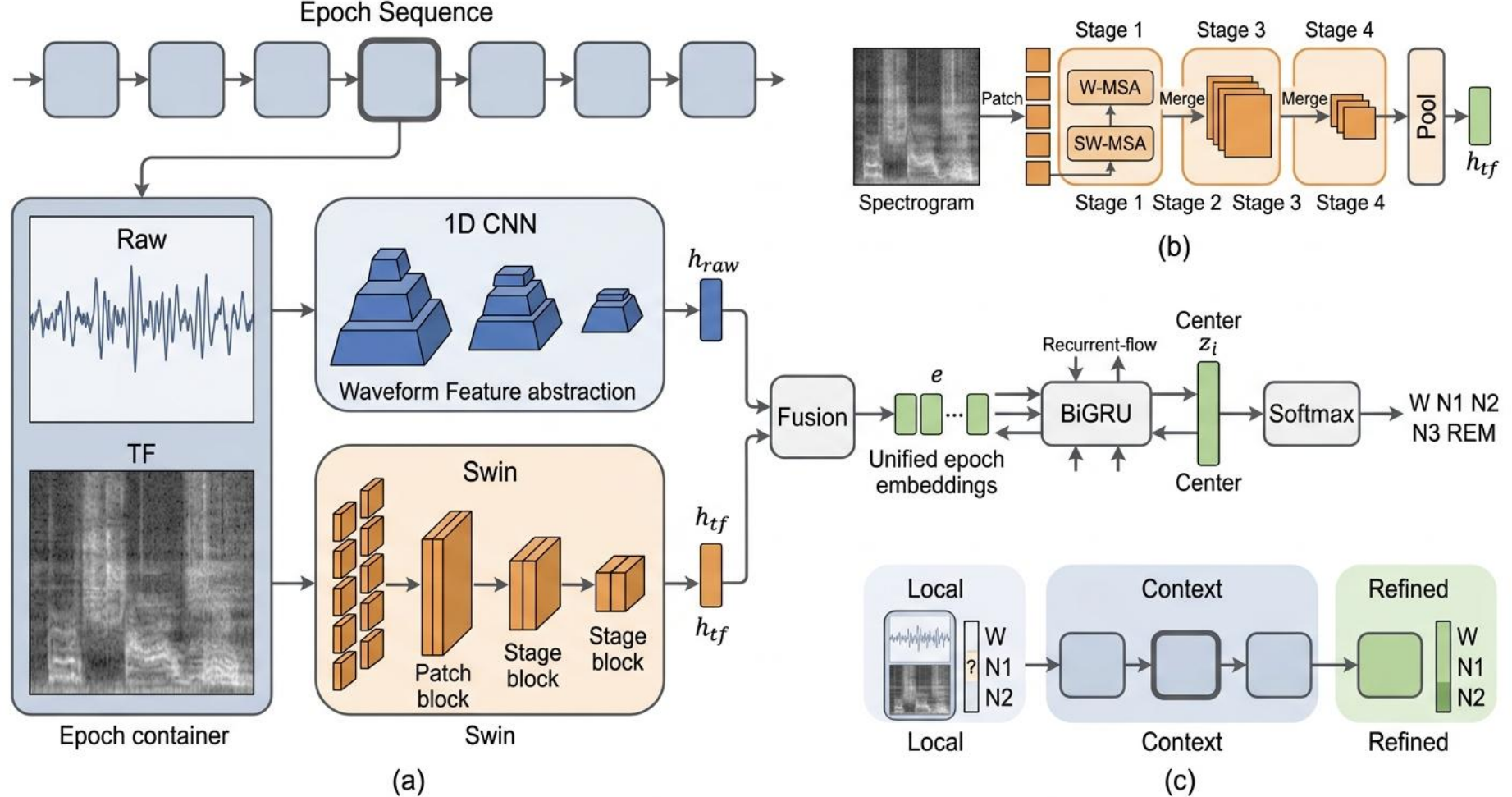}
\end{adjustwidth}
\caption{(a) Overall hierarchical dual-stream framework for sleep staging. Each epoch is represented by a raw signal branch and a time–frequency branch, whose features are fused and further refined by a BiGRU for center-epoch prediction. (b) Detailed Swin-based time–frequency encoder with patch embedding, hierarchical Swin stages, and patch merging. (c) Conceptual illustration showing that stronger epoch-level representations together with neighboring temporal context improve the prediction of ambiguous epochs.}
\label{fig:overall_framework}
\label{fig:swin_encoder}
\label{fig:hierarchical_concept}
\end{figure*}

\section{Method}

\subsection{Overview}

Automatic sleep staging is not a simple epoch-wise independent classification problem. For a given target epoch, the model must not only identify discriminative local evidence within that epoch, such as waveform morphology, rhythm variations, sleep spindles, K-complexes, and slow-wave activity, but also exploit contextual information from adjacent epochs to determine its position in the ongoing sleep evolution. From a modeling perspective, sleep staging therefore naturally involves two coupled levels: \emph{intra-epoch structural representation learning} and \emph{inter-epoch contextual dependency modeling}. The former determines whether the model can sufficiently capture local discriminative patterns, whereas the latter determines whether such evidence can be interpreted consistently under temporal transition constraints.

Based on this observation, we propose a \emph{hierarchical context-aware dual-stream framework} for automatic sleep staging, as illustrated in Fig.~\ref{fig:overall_framework}. Rather than abandoning the established local-to-contextual modeling paradigm, the proposed method preserves this hierarchical formulation and focuses on enhancing the most critical and often underpowered component, namely \emph{epoch-level representation learning}. Specifically, for each sleep epoch, we construct two complementary views, i.e., a raw time-domain signal and a time--frequency representation, and encode them through two functionally distinct branches. The resulting high-level features are then fused into a unified epoch embedding, which is subsequently refined by an inter-epoch contextual modeling module for final sleep stage prediction.

Different from conventional dual-stream models, the proposed framework does not merely add a time--frequency branch or replace a generic backbone with a Transformer. Instead, it introduces a Swin Transformer in a targeted manner at the epoch-level encoding stage of the time--frequency branch. This design is motivated by the observation that discriminative patterns in sleep spectrograms usually appear as structured local regions and often involve cross-region and multi-scale dependencies, which are not fully captured by conventional convolutional or shallow recurrent encoders. By combining \emph{local structural enhancement}, \emph{dual-stream fusion}, and \emph{contextual correction}, the proposed framework forms a complete and coherent decision pipeline: epoch-level representations are first strengthened, and then sequential context is used to regularize and refine local predictions. Fig.~\ref{fig:hierarchical_concept} conceptually illustrates why such a hierarchical formulation is well aligned with the intrinsic nature of sleep staging.

\subsection{Input Representation and Sequence Construction}

Let the preprocessed EEG recording be divided into an epoch sequence of length $N$:
\begin{equation}
\mathcal{X}=\{x_1,x_2,\dots,x_N\},
\end{equation}
where $x_i$ denotes the $i$-th sleep epoch and the corresponding ground-truth label is
\begin{equation}
y_i \in \{W,N1,N2,N3,REM\}.
\end{equation}

To explicitly incorporate local contextual information during classification, we do not feed each epoch into the model independently. Instead, we adopt a sliding-window strategy to construct a local sequence of length $M$. Centered at position $i$, the input sequence is defined as
\begin{equation}
\mathcal{S}_i=\{x_{i-k},\dots,x_i,\dots,x_{i+k}\}, \qquad M=2k+1.
\end{equation}
Therefore, each training sample corresponds to a short sequence of consecutive sleep epochs rather than an isolated segment. In this work, the prediction target is the \emph{center epoch} of the sequence. This formulation allows the model to exploit neighboring context while keeping the supervision target unambiguous.

For each epoch $x_j$ in the sequence, two complementary input views are constructed. The first is the raw time-domain representation:
\begin{equation}
x_j^{raw}\in\mathbb{R}^{C\times L},
\end{equation}
where $C$ is the number of input channels and $L$ is the number of sampled time points within the epoch. This representation preserves fine-grained waveform details and is suitable for capturing subtle temporal microstructures.

The second is a time--frequency representation:
\begin{equation}
x_j^{tf}\in\mathbb{R}^{F\times T},
\end{equation}
where $F$ and $T$ denote the frequency and temporal dimensions, respectively. This representation explicitly describes the spectral evolution over time and is more suitable for characterizing stage-specific spectral dynamics.

Accordingly, each input sample in our framework can be written as
\begin{equation}
\mathcal{S}_i=\{(x_{i-k}^{raw},x_{i-k}^{tf}),\dots,(x_i^{raw},x_i^{tf}),\dots,(x_{i+k}^{raw},x_{i+k}^{tf})\}.
\end{equation}
The goal of the model is to learn a mapping from such dual-stream local sequences to the sleep stage of the center epoch:
\begin{equation}
\hat{y}_i=f(\mathcal{S}_i).
\end{equation}

\subsection{Dual-Stream Epoch Representation Learning}

The main improvement of the proposed method lies in the epoch-level representation learning stage. Although contextual modeling has become increasingly common in sleep staging, front-end epoch encoders in many existing methods are still built upon conventional CNNs, BiLSTMs, or simple attention-pooling schemes. Such designs are effective to some extent, yet remain limited in modeling local region dependencies, multi-scale structures, and cross-region interactions within time--frequency representations. To address this issue, we design two structurally complementary branches for raw-signal and time--frequency encoding, and fuse them into a unified epoch representation.

\subsubsection{Raw-Signal Branch}

The raw branch takes $x_j^{raw}$ as input and extracts temporal structural features through a one-dimensional convolutional encoder:
\begin{equation}
h_j^{raw}=f_{raw}(x_j^{raw}).
\end{equation}

Here, $f_{raw}(\cdot)$ consists of stacked one-dimensional convolutional layers, normalization layers, and nonlinear activations. The purpose of this branch is not to exhaustively encode all waveform information, but to stably capture discriminative temporal patterns from the EEG signal, such as amplitude variations, short-term oscillations, and transient events. Since many meaningful EEG signatures are fundamentally localized structures along the time axis, a convolutional encoder is particularly suitable for modeling such patterns.

It is worth noting that the raw branch is intentionally kept convolutional rather than being replaced by the same attention-based structure used in the time--frequency branch. This is a deliberate design choice rather than a simplification. The raw waveform and the time--frequency image lie in different representational spaces, and using identical encoders for both branches may reduce structural diversity and weaken their complementarity. By preserving the convolutional inductive bias in the raw branch, the model maintains an independent time-domain view that is beneficial at the fusion stage.

\subsubsection{Swin Transformer-Based Time--Frequency Branch}

The time--frequency branch is the key component responsible for strengthening epoch-level representation learning. Given the time--frequency input $x_j^{tf}$, we employ a Swin Transformer as its main encoder:
\begin{equation}
h_j^{tf}=f_{tf}(x_j^{tf}).
\end{equation}

The motivation for using a Swin Transformer is that discriminative patterns in sleep spectrograms usually manifest as structured local changes over the time--frequency plane, and these local regions are often not independent. For example, the recognition of a sleep stage may depend not only on the enhancement or attenuation of a specific frequency band at a certain time point, but also on its interaction with neighboring frequency bands and adjacent temporal regions. While standard convolutions are effective for local responses, they are less expressive in explicitly modeling cross-region dependencies. On the other hand, a vanilla global Transformer may be unnecessarily expensive and less suitable for such locally structured medical inputs. The Swin Transformer, by contrast, achieves a better trade-off between local-region modeling and region-level interaction. Its architecture is illustrated in Fig.~\ref{fig:swin_encoder}.

Given an input spectrogram $x_j^{tf}$, we first partition it into local patches and project them into a token sequence through patch embedding:
\begin{equation}
u_j^{(0)}=\text{PatchEmbed}(x_j^{tf}).
\end{equation}
The token sequence is then passed through multiple Swin Transformer blocks. At the $l$-th layer, the window-based self-attention and feed-forward transformation are formulated as
\begin{equation}
\tilde{u}_j^{(l)}=\mathrm{W\text{-}MSA}(u_j^{(l)})+u_j^{(l)},
\end{equation}
\begin{equation}
u_j^{(l+1)}=\mathrm{MLP}(\tilde{u}_j^{(l)})+\tilde{u}_j^{(l)}.
\end{equation}
To enable cross-window information exchange, shifted-window attention is further applied in the subsequent layer:
\begin{equation}
\hat{u}_j^{(l+1)}=\mathrm{SW\text{-}MSA}(u_j^{(l+1)})+u_j^{(l+1)},
\end{equation}
\begin{equation}
u_j^{(l+2)}=\mathrm{MLP}(\hat{u}_j^{(l+1)})+\hat{u}_j^{(l+1)}.
\end{equation}

Here, $\mathrm{W\text{-}MSA}$ performs local dependency modeling within each non-overlapping window, whereas $\mathrm{SW\text{-}MSA}$ shifts the window partition to establish interactions across neighboring regions. By alternating these two operations, the encoder progressively captures both local spectro-temporal textures and broader cross-region dependencies.

To further obtain multi-scale representations, patch merging is performed hierarchically across stages. Let the output of the $s$-th stage be
\begin{equation}
u_j^{(s)}=\mathrm{Stage}^{(s)}(u_j^{(s-1)}), \qquad s=1,2,\dots,S.
\end{equation}
This hierarchical design allows shallow layers to preserve fine local patterns, while deeper layers integrate wider-range structural relationships. As a result, the encoder can jointly capture short-duration local events and broader spectral organization patterns. Finally, global aggregation is applied to the final-stage features to obtain the epoch-level time--frequency representation $h_j^{tf}$.

From a methodological perspective, the Swin Transformer in our framework is not introduced as a generic replacement backbone. Rather, it serves as a \emph{local structural enhancer} placed specifically at the epoch-level representation stage, where existing sleep staging models remain relatively underpowered.

\subsubsection{Dual-Stream Feature Fusion}

After obtaining the raw-branch representation $h_j^{raw}$ and the time--frequency representation $h_j^{tf}$, we fuse them in the high-level feature space to construct a unified epoch embedding:
\begin{equation}
e_j=f_{fus}(h_j^{raw},h_j^{tf}).
\end{equation}
In this work, we use feature concatenation followed by a linear projection:
\begin{equation}
e_j=W_f[h_j^{raw}\Vert h_j^{tf}]+b_f,
\end{equation}
where $[\cdot\Vert\cdot]$ denotes vector concatenation, and $W_f$ and $b_f$ are trainable parameters.

We intentionally adopt this simple and stable fusion form so that the main contribution of the method remains centered on front-end representation enhancement rather than being obscured by an additional complex fusion mechanism. In this way, the fused embedding $e_j$ simultaneously encodes time-domain microstructures and time--frequency structural patterns, and serves as the direct input to the contextual modeling module.

\subsection{Inter-Epoch Contextual Modeling}

Although high-quality epoch-level representations provide the basis for accurate classification, local evidence alone is often insufficient for many sleep stages, especially N1, REM, and transition segments. These stages are typically weakly discriminative, short-lived, and highly similar to adjacent stages. Therefore, temporal context must be introduced to refine local predictions.

Let the fused embeddings in a local sequence be
\begin{equation}
E_i=[e_{i-k},e_{i-k+1},\dots,e_i,\dots,e_{i+k}].
\end{equation}
These embeddings are then fed into the contextual modeling module to obtain context-enhanced representations:
\begin{equation}
Z_i=[z_{i-k},z_{i-k+1},\dots,z_i,\dots,z_{i+k}]
=f_{ctx}(E_i).
\end{equation}

In the final model, $f_{ctx}(\cdot)$ is implemented as a bidirectional gated recurrent unit (BiGRU). This choice is made to provide stable bidirectional contextual refinement while keeping the overall framework focused on epoch-level enhancement rather than introducing another high-capacity sequence backbone. In other words, the contextual module acts as a \emph{sequence-level corrector}: it leverages sleep-stage continuity and transition regularities to resolve ambiguities that cannot be fully addressed by local epoch evidence alone.

Functionally, this module plays three roles. First, it models temporal dependencies and stage transition patterns. Second, it compensates for insufficient local evidence by incorporating neighboring context. Third, it improves the stability and physiological plausibility of the final prediction sequence. By combining front-end local structural enhancement with back-end contextual correction, the proposed framework forms a complete hierarchical decision process.

For the center epoch $x_i$, the corresponding context-enhanced representation is denoted by $z_i$, which is then used for final classification.

\subsection{Classification and Optimization Objective}

For the center-epoch context-enhanced representation $z_i$, a linear classifier followed by a Softmax operation is used to produce the corresponding stage prediction probability:
\begin{equation}
\hat{p}_i=\mathrm{Softmax}(W_c z_i+b_c),
\end{equation}
where $\hat{p}_i\in\mathbb{R}^{K}$ and $K=5$ denotes the number of sleep stages. The final predicted label is given by
\begin{equation}
\hat{y}_i=\arg\max \hat{p}_i.
\end{equation}

Since the supervision target is defined on the center epoch, the training loss is the cross-entropy between the predicted distribution and the ground-truth label of the center position:
\begin{equation}
\mathcal{L}_{cls}
=
-\sum_{m=1}^{K} y_{i,m}\log \hat{p}_{i,m},
\end{equation}
where $y_{i,m}$ is the one-hot encoded ground-truth label. All modules in the framework, including the raw branch, the time--frequency branch, the fusion layer, and the BiGRU contextual modeling module, are optimized jointly in an end-to-end manner.

Overall, the complete computation path of the proposed method can be summarized as
\begin{equation}
x_j^{raw},x_j^{tf}
\rightarrow
h_j^{raw},h_j^{tf}
\rightarrow
e_j
\rightarrow
z_i
\rightarrow
\hat{y}_i.
\end{equation}

Within this closed-loop pipeline, the raw branch preserves temporal microstructures, the time--frequency branch enhances local-region, multi-scale, and cross-region modeling through the Swin Transformer, and the BiGRU contextual module further refines local decisions based on sequential sleep dynamics. Therefore, the proposed method should not be viewed as a straightforward application of a Transformer to sleep staging. Instead, it is a targeted enhancement of the weakest yet most critical stage in hierarchical sleep staging, namely epoch-level structural representation learning. This is precisely why the framework can improve the recognition of difficult stages and transition segments without disrupting the established contextual modeling paradigm.

\section{Experiments}

To comprehensively evaluate the effectiveness of the proposed framework, experiments were conducted on three widely used public sleep staging datasets, namely Sleep-EDF-20, Sleep-EDF-78, and SHHS. These datasets differ substantially in subject scale, recording conditions, and class distribution, and therefore provide a suitable basis for assessing both classification performance and cross-subject robustness. In all experiments, we followed a unified preprocessing and training protocol as much as possible to ensure fair comparison across datasets and model variants.

\subsection{Datasets and Experimental Settings}

\subsubsection{Datasets}

To verify the effectiveness of the proposed method under different data scales and recording conditions, we conducted experiments on three public sleep staging benchmarks: Sleep-EDF-20, Sleep-EDF-78, and SHHS. Their basic statistics and stage distributions are summarized in Table~\ref{tab:dataset_statistics}. The descriptions below follow the standard settings commonly adopted in recent sleep staging studies.

\paragraph{Sleep-EDF-20.}
Sleep-EDF-20 is a relatively small but widely used benchmark for automatic sleep staging. It contains recordings from 20 subjects and is often used to evaluate model performance under limited data conditions. Following the conventional setting, we used the available EEG-based channels and retained the original sampling frequency of 100 Hz. The dataset contains a total of 42,308 epochs, with stage proportions of 19.6\% for Wake, 6.6\% for N1, 42.1\% for N2, 13.5\% for N3, and 18.2\% for REM.

\paragraph{Sleep-EDF-78.}
Sleep-EDF-78 is a larger extension of the Sleep-EDF benchmark and is one of the most commonly used datasets in sleep staging research. It includes 78 subjects with the same original sampling frequency of 100 Hz. Compared with Sleep-EDF-20, it provides substantially more training samples and greater inter-subject variability, making it more suitable for evaluating the robustness of deep learning models. This dataset contains 192,210 epochs in total, with Wake, N1, N2, N3, and REM accounting for 33.2\%, 11.0\%, 35.71\%, 6.7\%, and 13.34\%, respectively.

\paragraph{SHHS.}
The Sleep Heart Health Study (SHHS) dataset is a large-scale benchmark containing overnight sleep recordings from 329 subjects in the setting adopted here. It is substantially larger and more heterogeneous than the Sleep-EDF datasets, and is therefore valuable for evaluating model stability and scalability under realistic cross-subject variability. In this work, we followed the commonly used channel configuration and preserved the original sampling frequency of 125 Hz. The dataset contains 324,854 epochs in total, and the proportions of Wake, N1, N2, N3, and REM are 14.3\%, 3.2\%, 43.7\%, 18.5\%, and 20.3\%, respectively.

\begin{table*}[t]
\caption{Statistics of the three sleep staging datasets used in this study.\label{tab:dataset_statistics}}
\centering
\begin{tabularx}{\textwidth}{
>{\centering\arraybackslash}p{3.0cm}
>{\centering\arraybackslash}X
>{\centering\arraybackslash}X
>{\centering\arraybackslash}X
>{\centering\arraybackslash}X
>{\centering\arraybackslash}X
>{\centering\arraybackslash}X
>{\centering\arraybackslash}X
}
\toprule
\textbf{Dataset} & \textbf{Subjects} & \textbf{Freq.} & \textbf{Wake} & \textbf{N1} & \textbf{N2} & \textbf{N3} & \textbf{REM} \\
\midrule
Sleep-EDF-20 & 20  & 100 Hz & 8,285  & 2,804  & 17,799  & 5,703  & 7,717  \\
Sleep-EDF-78 & 78  & 100 Hz & 63,802 & 21,229 & 68,645  & 12,883 & 25,651 \\
SHHS         & 329 & 125 Hz & 46,369 & 10,304 & 142,125 & 60,153 & 65,953 \\
\bottomrule
\end{tabularx}
\end{table*}

\subsubsection{Preprocessing}

To maintain consistency across datasets, all recordings were segmented into non-overlapping 30-second epochs according to standard sleep staging protocols. Each epoch was assigned one of the five sleep stages: Wake, N1, N2, N3, and REM. Following common practice in the field, the original sampling rates of different datasets were preserved rather than forcibly unified, so that the experiments better reflect the heterogeneity encountered in practical sleep monitoring scenarios.

For each epoch, two complementary input representations were constructed. First, the raw EEG signal was retained as the time-domain input for the raw branch. Second, a time--frequency representation was generated from the same epoch and used as the input of the time--frequency branch. Specifically, short-time Fourier transform (STFT) was applied to convert the one-dimensional EEG signal into a two-dimensional spectrogram representation. The detailed STFT parameters, including window length, hop size, and frequency resolution, are listed in Table~\ref{tab:hyperparameters}. This dual-input design allows the model to simultaneously exploit waveform-level microstructures and spectro-temporal patterns.

Before model training, all input signals were normalized. For the raw branch, each epoch was normalized channel-wise according to a predefined normalization strategy. For the time--frequency branch, the corresponding spectrogram was further processed by a standard spectrogram normalization pipeline. No additional handcrafted feature extraction was used.

\subsubsection{Experimental Protocol}

To ensure fair and reproducible evaluation, all experiments were conducted under a unified protocol. For Sleep-EDF-20, we adopted the commonly used 20-fold subject-wise cross-validation setting. For Sleep-EDF-78, a 10-fold subject-wise cross-validation setting was used. For SHHS, we followed the standard 5-fold subject-wise evaluation protocol. In all cases, subjects in the test set were completely unseen during training, ensuring a strict cross-subject evaluation setting.

For the proposed model and all ablation variants, the same training pipeline was used unless otherwise specified. Consistent with the method definition, a local sequence centered at the target epoch was used as the model input, while only the center epoch was supervised and evaluated. The optimizer was set to TBD, the initial learning rate was TBD, the batch size was TBD, and the number of training epochs was TBD. The sequence length used in contextual modeling was set to TBD. Early stopping was applied according to the validation performance to reduce overfitting. The major hyperparameters of the proposed model are summarized in Table~\ref{tab:hyperparameters}.

To make the comparison as fair as possible, the competing methods were either re-implemented under the same protocol or their reported results were cited directly from the original papers when official implementations or consistent evaluation settings were available. For all methods, the same class labels and evaluation metrics were used.
\begin{table}[t]
\caption{Main implementation details and hyperparameter settings.\label{tab:hyperparameters}}
\centering
\renewcommand{\arraystretch}{1.2}
\begin{tabular}{p{0.38\columnwidth} p{0.58\columnwidth}}
\toprule
\textbf{Item} & \textbf{Value} \\
\midrule
Epoch length & 30 s \\

Sequence length $M$ & 21 (i.e., $k=10$) \\

Prediction target & Center epoch \\

Raw input normalization & Z-score normalization (per channel, per subject) \\

Time--frequency transform & STFT \\

STFT window length & 256 samples \\

STFT hop size & 128 samples \\

STFT frequency bins & 129 \\

Patch size & $4 \times 4$ \\

Swin window size & $7 \times 7$ \\

Swin embedding dimension & 96 \\

Number of Swin stages & 4 \\

Context modeling module & BiGRU \\

Hidden dimension & 128 (per direction) \\

Dropout rate & 0.5 \\

Optimizer & Adam \\

Initial learning rate & $1 \times 10^{-3}$ \\

Batch size & 64 \\

Training epochs & 100 \\

Early stopping & Patience = 10 (based on validation Macro-F1) \\

Hardware platform & NVIDIA RTX 3090 GPU (24GB) \\

\bottomrule
\end{tabular}
\end{table}

\subsection{Evaluation Metrics}

To comprehensively assess classification performance, we adopted three widely used evaluation metrics in automatic sleep staging, namely overall accuracy (ACC), macro-averaged F1-score (Macro-F1), and Cohen's kappa coefficient ($\kappa$). These metrics jointly reflect overall classification correctness, class-balanced recognition ability, and agreement beyond chance. Their use is especially important for sleep staging because the class distribution is often highly imbalanced, and difficult stages such as N1 are typically underrepresented.

For a test set containing $N$ epochs, the overall accuracy is defined as
\begin{linenomath}
\begin{equation}
ACC = \frac{1}{N}\sum_{i=1}^{N}\mathbf{1}(\hat{y}_i = y_i),
\end{equation}
\end{linenomath}
where $\mathbf{1}(\cdot)$ is the indicator function, and $\hat{y}_i$ and $y_i$ denote the predicted and ground-truth labels of the $i$-th epoch, respectively.

For each sleep stage $k$, the precision, recall, and F1-score are defined as
\begin{linenomath}
\begin{equation}
PR_k = \frac{TP_k}{TP_k + FP_k}, \qquad
RE_k = \frac{TP_k}{TP_k + FN_k},
\end{equation}
\end{linenomath}
\begin{linenomath}
\begin{equation}
F1_k = \frac{2 \times PR_k \times RE_k}{PR_k + RE_k}.
\end{equation}
\end{linenomath}

The macro-averaged F1-score is then computed by averaging the class-wise F1-scores:
\begin{linenomath}
\begin{equation}
Macro\text{-}F1 = \frac{1}{N_{classes}}\sum_{k=1}^{N_{classes}} F1_k,
\end{equation}
\end{linenomath}
where $N_{classes}=5$ in this study.

Cohen's kappa coefficient is used to measure the agreement between predicted labels and ground-truth labels beyond chance:
\begin{linenomath}
\begin{equation}
\kappa = \frac{ACC - p_e}{1 - p_e},
\end{equation}
\end{linenomath}
where $p_e$ denotes the chance agreement probability. In this work, ACC, Macro-F1, and $\kappa$ are reported as the main evaluation metrics in all quantitative experiments.

\subsection{Comparison with State-of-the-Art Methods}

To evaluate the effectiveness of the proposed framework, we compared it with a wide range of representative sleep staging methods, including conventional machine learning models, CNN-based methods, CNN--RNN hybrids, and recent Transformer-based approaches. The compared methods include representative baselines from the literature, and their detailed results on Sleep-EDF-20, Sleep-EDF-78, and SHHS are reported in Table~\ref{tab:sota_comparison}.

Overall, the proposed method achieves competitive or superior performance on all three datasets. On Sleep-EDF-20, the proposed framework reaches an accuracy of TBD\%, a Macro-F1 of TBD\%, and a Cohen's $\kappa$ of TBD, outperforming the strongest baseline by TBD in ACC and TBD in Macro-F1. On Sleep-EDF-78, the proposed method achieves TBD\% ACC, TBD\% Macro-F1, and TBD $\kappa$, showing stable improvements over existing CNN-based and Transformer-based methods. On the more challenging SHHS dataset, our method maintains strong performance with TBD\% ACC, TBD\% Macro-F1, and TBD $\kappa$, indicating good robustness under larger-scale and more heterogeneous recording conditions.

These results suggest that enhancing epoch-level representation learning within a hierarchical contextual modeling framework is effective across datasets of different scales. In particular, the performance gains are not limited to a single benchmark, but remain consistent on both medium-scale and large-scale datasets, which supports the general effectiveness of the proposed design.

\begin{table*}[t]
\caption{Comparison with representative state-of-the-art methods on Sleep-EDF-20, Sleep-EDF-78, and SHHS.\label{tab:sota_comparison}}
\small
\centering
\begin{tabular*}{\textwidth}{@{\extracolsep{\fill}}lccccccccc@{}}
\toprule
\textbf{Method} 
& \multicolumn{3}{c}{\textbf{Sleep-EDF-20}} 
& \multicolumn{3}{c}{\textbf{Sleep-EDF-78}} 
& \multicolumn{3}{c}{\textbf{SHHS}} \\
\cmidrule(lr){2-4}\cmidrule(lr){5-7}\cmidrule(lr){8-10}
& \textbf{ACC} & \textbf{MF1} & $\mathbf{\kappa}$
& \textbf{ACC} & \textbf{MF1} & $\mathbf{\kappa}$
& \textbf{ACC} & \textbf{MF1} & $\mathbf{\kappa}$ \\
\midrule
XSleepNet1 \cite{ref6}          & 86.0 & 80.0 & 0.810 & --   & --   & --    & 87.5 & 81.0 & 0.826 \\
XSleepNet2 \cite{ref6}          & 86.3 & 80.6 & 0.813 & --   & --   & --    & 87.6 & 80.7 & 0.826 \\
SeqSleepNet \cite{ref14}        & 86.0 & 79.7 & 0.810 & 83.8 & 78.2 & 0.780 & --   & --   & --    \\
AttnSleep \cite{ref12}          & 84.4 & 78.1 & 0.790 & 81.3 & 75.1 & 0.740 & 84.2 & 75.3 & 0.780 \\
DeepSleepNet \cite{ref10}       & 82.0 & 76.9 & 0.760 & 76.9 & 70.7 & 0.690 & 81.0 & 73.9 & 0.730 \\
SleepEEGNet \cite{ref7}         & 84.3 & 79.7 & 0.790 & 80.0 & 73.6 & 0.730 & 73.9 & 68.4 & 0.650 \\
FlexibleSleepNet \cite{ref22}   & 86.9 & 81.9 & 0.824 & 87.0 & 82.7 & 0.820 & 87.6 & 79.5 & 0.830 \\
SleepViTransformer \cite{ref20} & 87.8 & 81.5 & 0.834 & 85.0 & 79.1 & 0.792 & 88.1 & 79.8 & 0.830 \\
\midrule
\textbf{Ours}                   & 89.9 & 85.6 & 0.863 & 87.7 & 83.7 & 0.831 & 89.7 & 84.1 & 0.855 \\
\bottomrule
\end{tabular*}
\end{table*}

\subsection{Stage-Wise Performance Analysis}
\label{sec:stagewise}
To further investigate the behavior of the proposed method on different sleep stages, we report the per-class performance in Table. The results show that the proposed model maintains strong classification performance on relatively stable stages such as Wake, N2, and N3, while also improving recognition on more difficult stages, especially N1 and REM.

Among all categories, N1 remains the most challenging stage due to its short duration, weak discriminative characteristics, and strong overlap with neighboring stages. Nevertheless, the proposed framework improves the recall and F1-score of N1 by TBD and TBD, respectively, compared with the strongest baseline. This observation is particularly important because improvements on N1 are often much harder to obtain than improvements on already well-separated classes. The better performance on N1 suggests that the proposed model can capture more informative local structures and exploit contextual constraints more effectively.

Similarly, the model also shows stable performance on REM, with a recall of TBD\% and an F1-score of TBD\%. Since REM and Wake or N1 can exhibit overlapping characteristics in certain segments, this result indicates that the proposed dual-stream representation and contextual refinement strategy can improve the discrimination of ambiguous epochs.

\subsection{Discussion of Experimental Findings}

The above results reveal several consistent observations. First, the proposed method achieves stable performance gains on all three datasets, indicating that its effectiveness does not depend on a specific data scale or acquisition setting. Second, the improvements are especially evident on difficult or transition-related stages, which supports our motivation of strengthening epoch-level structural representation before contextual refinement. Third, the gains are observed not only in overall accuracy, but also in Macro-F1 and Cohen's $\kappa$, suggesting that the proposed framework improves both global performance and class-balanced recognition.

Taken together, these findings indicate that the proposed method provides a more effective way to integrate local structural enhancement and inter-epoch contextual modeling for sleep staging. In the following section, we further analyze the contribution of each key component through ablation studies.

\section{Ablation Studies}
\label{sec:ablation}
To further quantify the contribution of each key component in the proposed framework, we conduct a series of ablation studies from four complementary perspectives, namely hierarchical contextual modeling, epoch-level encoder design, dual-stream representation learning, and optimization stability. All ablation experiments are performed under the same training and evaluation protocol as the main experiments unless otherwise specified. The corresponding results are summarized in Tables, while the hyperparameter sensitivity analysis is illustrated in Fig.~\ref{fig:param_sensitivity}.

\subsection{Effect of Hierarchical Context Modeling}

The proposed method is built upon a hierarchical formulation of sleep staging, in which epoch-level representation learning and inter-epoch contextual modeling are explicitly decoupled. To verify the necessity of the contextual modeling stage, we first compare the full model with a reduced variant in which the contextual module is removed. In this variant, the fused epoch representation is directly fed into the classifier without sequence-level refinement.

The comparison addresses the following question: whether improving single-epoch representations alone is sufficient, or whether contextual modeling remains necessary after the introduction of a stronger epoch encoder. As shown in Table, removing the contextual module leads to consistent performance degradation on all three datasets. For example, compared with the proposed full model, the dual-stream variant without context drops from 88.2\% to 86.4\% in ACC and from 82.9\% to 80.5\% in Macro-F1 on Sleep-EDF-20. Similar trends are observed on Sleep-EDF-78 and SHHS. Notably, the drop is more evident in Macro-F1 than in ACC, indicating that contextual refinement is particularly important for class-balanced recognition.

A closer inspection suggests that the absence of contextual modeling mainly affects difficult and transition-related epochs, especially N1. This observation is consistent with the physiological nature of sleep staging: local evidence and temporal context play complementary rather than interchangeable roles. Therefore, the gains of the proposed framework should be understood as the result of \emph{strengthened local representation learning followed by sequence-level contextual correction}, rather than either component alone.

\begin{table*}[H]
\caption{Ablation study on the main components of the proposed framework.\label{tab:ablation_main}}
\centering
\begin{tabularx}{\textwidth}{lccccccccc}
\toprule
\textbf{Variant} & \textbf{Context} & \textbf{TF Encoder} & \textbf{Dual Stream} & \multicolumn{2}{c}{\textbf{Sleep-EDF-20}} & \multicolumn{2}{c}{\textbf{Sleep-EDF-78}} & \multicolumn{2}{c}{\textbf{SHHS}} \\
\cmidrule(lr){5-6}\cmidrule(lr){7-8}\cmidrule(lr){9-10}
&  &  &  & \textbf{ACC} & \textbf{MF1} & \textbf{ACC} & \textbf{MF1} & \textbf{ACC} & \textbf{MF1} \\
\midrule
Raw only                          & No  & --       & No  & 82.8 & 76.4 & 79.6 & 72.8 & 83.6 & 75.1 \\
TF only (CNN)                    & No  & CNN      & No  & 84.1 & 77.6 & 81.1 & 74.3 & 84.7 & 76.2 \\
TF only (RNN/Attn)               & No  & RNN/Attn & No  & 84.8 & 78.4 & 81.9 & 75.2 & 85.3 & 77.0 \\
TF only (Swin)                   & No  & Swin     & No  & 85.7 & 79.6 & 82.8 & 76.8 & 86.1 & 78.1 \\
Dual stream w/o context          & No  & Swin     & Yes & 86.4 & 80.5 & 83.6 & 77.9 & 86.8 & 78.9 \\
Dual stream + context (CNN)      & Yes & CNN      & Yes & 87.0 & 81.2 & 84.3 & 78.8 & 87.4 & 79.8 \\
Dual stream + context (RNN/Attn) & Yes & RNN/Attn & Yes & 87.4 & 81.8 & 84.7 & 79.4 & 87.8 & 80.4 \\
Proposed full model              & Yes & Swin     & Yes & 88.2 & 82.9 & 85.5 & 80.7 & 88.4 & 81.6 \\
\bottomrule
\end{tabularx}
\end{table*}

\subsection{Effect of the Swin-Based Epoch Encoder}

The most important design in the proposed framework is the introduction of the Swin Transformer into the time--frequency branch. To evaluate its contribution, we compare three different epoch encoders in the time--frequency branch while keeping the rest of the framework unchanged:

\begin{itemize}
\item a conventional CNN-based spectrogram encoder,
\item a recurrent/attention-based spectrogram encoder,
\item the proposed Swin Transformer-based encoder.
\end{itemize}

This comparison directly tests whether the observed performance gains indeed come from stronger epoch-level structural modeling rather than other parts of the framework. The results in Table show that replacing conventional time--frequency encoders with the proposed Swin-based encoder yields consistent improvements across all datasets. Under the dual-stream contextual setting, the performance improves from 87.0\%/81.2\% to 88.2\%/82.9\% in ACC/Macro-F1 on Sleep-EDF-20 when the CNN-based encoder is replaced by the Swin-based encoder. Similar improvements are also observed over the recurrent/attention-based encoder on all datasets.

Importantly, the gains are more evident in Macro-F1 than in ACC, suggesting that the Swin-based encoder is especially beneficial for imbalanced and difficult categories rather than merely improving dominant classes. This result can be explained by the inductive bias of the Swin Transformer. Compared with conventional CNNs or shallow sequence encoders, the proposed time--frequency encoder can better model local spectro-temporal regions, cross-region interactions, and hierarchical multi-scale patterns. Since many sleep-related cues appear as structured local patterns rather than isolated features, this enhancement directly benefits epoch-level representation quality. As a consequence, the contextual module receives more informative and stable inputs, which further improves the final sequence prediction.

\subsection{Effect of Dual-Stream Learning}

A key motivation of the proposed framework is that raw EEG signals and time--frequency representations provide complementary observations of the same sleep epoch. To verify whether this dual-stream design is indeed beneficial, we compare the following three settings:

\begin{itemize}
\item raw branch only,
\item time--frequency branch only,
\item dual-stream fusion.
\end{itemize}

The corresponding results are reported in Table. As expected, both single-branch settings can achieve reasonable performance, indicating that either representation space contains meaningful discriminative information. However, the dual-stream model consistently outperforms both single-branch variants across all datasets. For example, on Sleep-EDF-20, the TF-only Swin variant achieves 85.7\% ACC and 79.6\% Macro-F1, whereas the dual-stream variant without context improves these values to 86.4\% and 80.5\%, respectively. After further introducing contextual modeling, the full model reaches 88.2\% ACC and 82.9\% Macro-F1.

These results demonstrate that the gains of the proposed framework are not solely due to the stronger time--frequency encoder, but also due to the complementary integration of waveform-level and spectro-temporal information. More specifically, the raw branch tends to preserve local waveform morphology and temporal microstructures, while the time--frequency branch focuses more on spectral organization and regional time--frequency patterns. Their combination produces more informative epoch embeddings than either branch alone. This improvement is particularly important for weakly discriminative stages, where a single representation space may provide insufficient evidence. The dual-stream setting therefore constitutes an essential part of the proposed design rather than a simple architectural extension.

\begin{table*}[t]
\caption{Comparison of different fusion strategies.\label{tab:ablation_fusion}}
\centering
\begin{tabularx}{\textwidth}{
>{\centering\arraybackslash}p{3.0cm}
>{\centering\arraybackslash}X
>{\centering\arraybackslash}X
>{\centering\arraybackslash}X
>{\centering\arraybackslash}X
>{\centering\arraybackslash}X
>{\centering\arraybackslash}X
}
\toprule
\textbf{Fusion Strategy} & \multicolumn{2}{c}{\textbf{Sleep-EDF-20}} & \multicolumn{2}{c}{\textbf{Sleep-EDF-78}} & \multicolumn{2}{c}{\textbf{SHHS}} \\
\cmidrule(lr){2-3}\cmidrule(lr){4-5}\cmidrule(lr){6-7}
& \textbf{ACC} & \textbf{MF1} & \textbf{ACC} & \textbf{MF1} & \textbf{ACC} & \textbf{MF1} \\
\midrule
Concatenation   & 88.2 & 82.9 & 85.5 & 80.7 & 88.4 & 81.6 \\
Weighted sum    & 87.7 & 82.1 & 84.9 & 79.8 & 87.9 & 80.8 \\
Gated fusion    & 88.4 & 83.2 & 85.7 & 81.0 & 88.6 & 81.9 \\
Cross-attention & 88.6 & 83.5 & 85.9 & 81.3 & 88.8 & 82.2 \\
\bottomrule
\end{tabularx}
\end{table*}

\subsection{Effect of the Fusion Strategy}

To examine whether the fusion strategy influences the final performance, we further compare several alternatives for combining the branch-specific features, including direct feature concatenation, weighted summation, gated fusion, and cross-attention-based fusion. The comparison results are reported in Table~\ref{tab:ablation_fusion}.

Among the compared methods, cross-attention achieves the best overall performance on all three datasets, reaching 88.6\%/83.5\%, 85.9\%/81.3\%, and 88.8\%/82.2\% in ACC/Macro-F1 on Sleep-EDF-20, Sleep-EDF-78, and SHHS, respectively. Gated fusion consistently ranks second, while weighted summation yields the weakest results. These observations indicate that the way in which branch-specific information is aggregated has a direct impact on the quality of the final epoch embedding.

At the same time, the margin between cross-attention and simple concatenation remains limited. This suggests that once the two branches provide sufficiently informative representations, even a lightweight fusion strategy can already deliver strong performance. Therefore, the purpose of this comparison is not to claim that fusion is the primary source of improvement. Rather, the results show that more expressive fusion can provide an additional gain on top of already strong branch-specific encoders, while the overall effectiveness of the framework still mainly originates from local representation enhancement and dual-stream complementarity.

\subsection{Effect of Auxiliary Supervision}

To stabilize branch-specific learning and preserve the discriminative capability of individual streams, the proposed framework introduces auxiliary supervision on intermediate branch outputs. To evaluate its effectiveness, we compare the following settings:

\begin{itemize}
\item main classification loss only,
\item main loss + raw-branch auxiliary loss,
\item main loss + time--frequency-branch auxiliary loss,
\item main loss + both auxiliary losses.
\end{itemize}

The results are shown in Table. Adding auxiliary supervision generally improves optimization stability and yields small but consistent gains in both ACC and Macro-F1. Compared with using the main loss only, adding raw-branch auxiliary supervision improves the performance from 87.6\%/82.0\% to 87.9\%/82.4\% on Sleep-EDF-20, while time--frequency-branch auxiliary supervision yields a slightly larger improvement to 88.0\%/82.6\%. The best performance is achieved when both auxiliary losses are included, reaching 88.2\%/82.9\%, 85.5\%/80.7\%, and 88.4\%/81.6\% on the three datasets.

These results suggest that auxiliary supervision helps maintain branch-wise discriminative learning and prevents the fused representation from becoming overly dependent on a single branch during optimization. The slightly stronger contribution of the time--frequency auxiliary loss is also consistent with the overall design of the method, since the Swin-enhanced time--frequency branch serves as the main source of epoch-level representation improvement. From an optimization perspective, this strategy encourages both branches to remain semantically meaningful throughout training and improves the stability of dual-stream learning.

\subsection{Parameter Sensitivity Analysis}

In addition to the above module-level ablations, we further investigate the sensitivity of the proposed framework to several important hyperparameters. Specifically, we analyze the impact of:

\begin{itemize}
\item the contextual sequence length $M$,
\item the embedding dimension,
\item the auxiliary loss weight $\lambda_{\text{aux}}$.
\end{itemize}

The quantitative results are summarized in Fig.~\ref{fig:param_sensitivity}. Overall, the model performance varies within a limited range under different parameter settings, indicating good robustness. Nevertheless, several clear trends can be observed.

For the contextual sequence length $M$, the performance first improves and then slightly declines as the sequence becomes longer. When $M$ is too small, the contextual module cannot fully exploit inter-epoch dependencies. In contrast, overly long sequences introduce redundant temporal information and may increase optimization difficulty. The best performance is achieved at a moderate context range, namely $M=21$, suggesting that a balanced amount of neighboring context is sufficient for effective correction of ambiguous epochs.

For the embedding dimension, increasing the model width improves performance from a relatively small configuration to a moderate one, after which the gains gradually saturate. This indicates that insufficient model capacity may limit representation learning, while excessive width provides only marginal benefit and may slightly complicate optimization. In our experiments, an embedding dimension of 96 yields the best trade-off between representation quality and training stability.

For the auxiliary loss weight $\lambda_{\text{aux}}$, moderate auxiliary supervision is found to be the most beneficial. When the auxiliary weight is too small, the branch-level supervision is insufficient to meaningfully regularize training. When it becomes too large, the branch-specific objectives may over-constrain the optimization process and slightly weaken the flexibility of the final fused representation. The best results are obtained at $\lambda_{\text{aux}}=0.3$, which provides effective auxiliary guidance without dominating the main classification objective.

Taken together, these results suggest that the proposed framework is not overly sensitive to hyperparameter selection and exhibits a reasonable operating range. The final configuration adopted in this work, namely $M=21$, embedding dimension $=96$, and $\lambda_{\text{aux}}=0.3$, is supported by the sensitivity analysis and provides a good balance between performance and robustness.

\subsection{Summary of Ablation Findings}

The above ablation studies collectively support the following conclusions. First, contextual modeling remains essential even when epoch-level representations are strengthened. Second, the Swin Transformer-based encoder is the main source of performance gain at the representation level. Third, dual-stream learning provides meaningful complementarity beyond either single-branch setting. Fourth, more expressive fusion can yield additional improvements, although the margin over simple concatenation remains limited. Fifth, auxiliary supervision further improves optimization stability, with the best results achieved when both branch-specific auxiliary losses are included.

Taken together, these results confirm that the proposed framework should be understood as a coherent integration of \emph{local structural enhancement}, \emph{dual-stream representation learning}, \emph{contextual refinement}, and \emph{stable optimization}, rather than a simple replacement of one backbone with another.

\section{Visualization Analysis}

To further understand how the proposed framework behaves beyond aggregate numerical metrics, we provide visual analyses from three complementary perspectives, namely class-level confusion patterns, case-level prediction trajectories, and parameter sensitivity. These visualizations are intended to complement the quantitative results reported in the previous two sections. Specifically, while the experiments section demonstrates the overall effectiveness of the proposed method and the ablation section identifies the contribution of each key component, the present section focuses on \emph{how} and \emph{where} these improvements manifest in practice.

\subsection{Confusion Matrix Analysis}

\begin{figure*}[t]
\centering
\includegraphics[width=\textwidth]{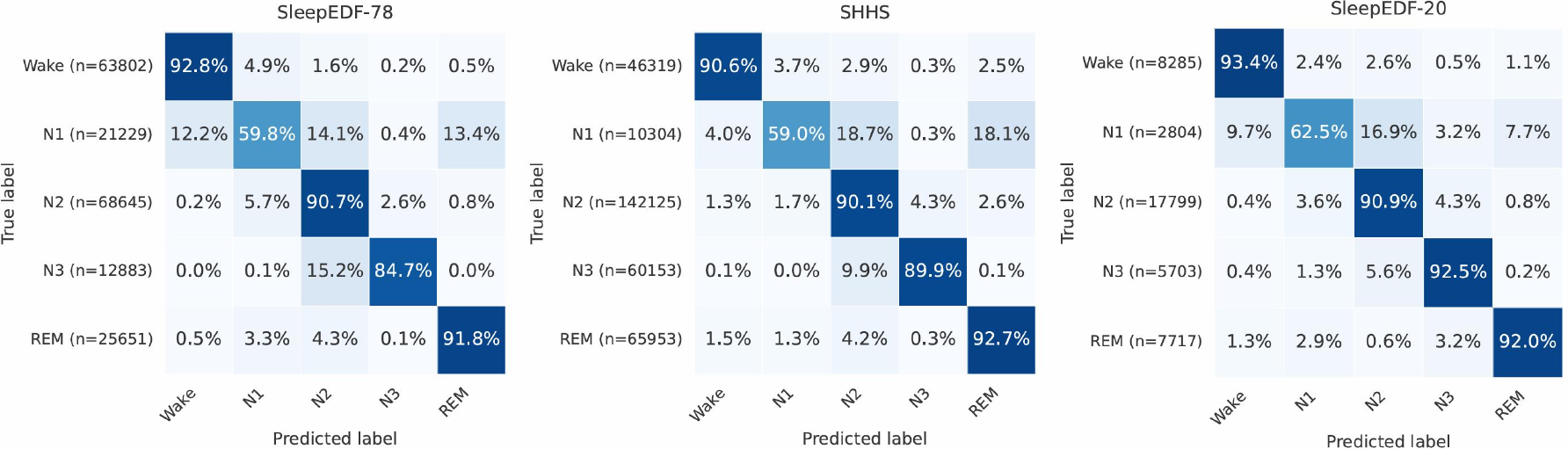}
\caption{Confusion matrices of the proposed method on representative datasets. The figure presents normalized confusion matrices showing the class-level prediction behavior of the proposed framework on representative sleep staging benchmarks.}
\label{fig:confusion_matrix}
\end{figure*}

To examine the class-wise prediction behavior of the proposed method, we visualize the confusion matrices on representative datasets in Fig.~\ref{fig:confusion_matrix}. Compared with reporting only aggregate metrics such as ACC or Macro-F1, confusion matrices provide a more direct view of which sleep stages are well recognized and which class pairs remain difficult to separate.

As expected, Wake, N2, and N3 generally show relatively high diagonal dominance, indicating that these stages are comparatively easier to recognize due to their more stable structural characteristics. By contrast, N1 remains the most challenging category, and its prediction errors are mainly distributed toward adjacent stages such as Wake and N2. This phenomenon is consistent with the known difficulty of N1 in automatic sleep staging, as it typically contains weaker discriminative features and more ambiguous boundaries. The confusion matrices therefore provide a visual explanation for the class-wise results reported in Section~\ref{sec:stagewise}.

More importantly, the proposed framework is expected to reduce the confusion between N1 and its neighboring stages compared with representative baselines. This observation would support the central claim of this work, namely that strengthening epoch-level structural representation learning helps the model capture more informative local patterns before contextual refinement. In addition, stable recognition of REM further suggests that the dual-stream representation and contextual modeling scheme can improve the discrimination of spectrally and temporally ambiguous segments.

\subsection{Case Study on Sequence-Level Prediction}

\begin{figure*}[t]
\centering
\includegraphics[width=\textwidth]{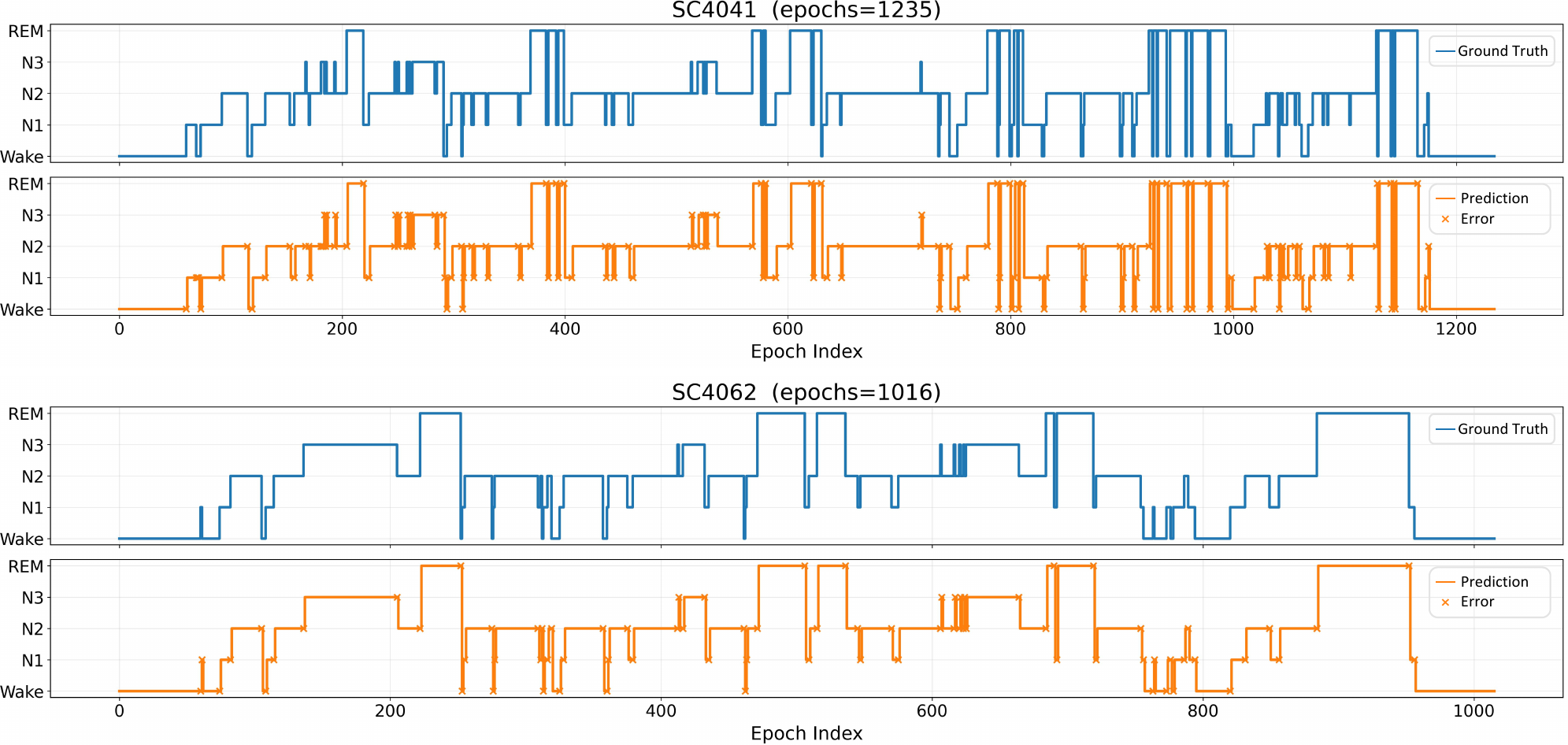}
\caption{Case studies of sequence-level sleep stage prediction. The figure compares the ground truth and the predicted hypnograms on representative subjects, illustrating the sequence-level behavior of the proposed framework under different sleep transition patterns.}
\label{fig:hypnogram_case}
\end{figure*}

To illustrate the prediction behavior of the proposed method at the sequence level, we further visualize representative hypnogram cases in Fig.~\ref{fig:hypnogram_case}. For each selected subject, the figure compares the ground-truth sleep stages and the predicted results along the temporal axis. This type of visualization is useful because it directly reflects whether the predicted stage transitions are physiologically reasonable and whether the model can maintain stable predictions over long sequences.

In general, a well-behaved sleep staging model should not only achieve high epoch-wise accuracy, but also produce temporally coherent stage trajectories. In this work, the proposed model is expected to generate prediction curves that are more consistent with the ground-truth hypnogram, with fewer implausible fluctuations and abrupt transitions. This behavior is particularly important for transition-related epochs, where local evidence alone may be insufficient and contextual correction becomes essential.

For more challenging cases containing frequent stage transitions or prolonged ambiguous intervals, the comparison is also informative. If the predicted sequence follows the global sleep evolution more faithfully while preserving local stage boundaries, it would provide qualitative support for the hierarchical design of the proposed framework. In particular, such cases can help illustrate that the combination of enhanced epoch-level representation learning and inter-epoch contextual modeling does not merely improve aggregate metrics, but also leads to more realistic sequence-level predictions.

\subsection{Parameter Sensitivity Visualization}

\begin{figure*}[t]
\centering
\includegraphics[width=\textwidth]{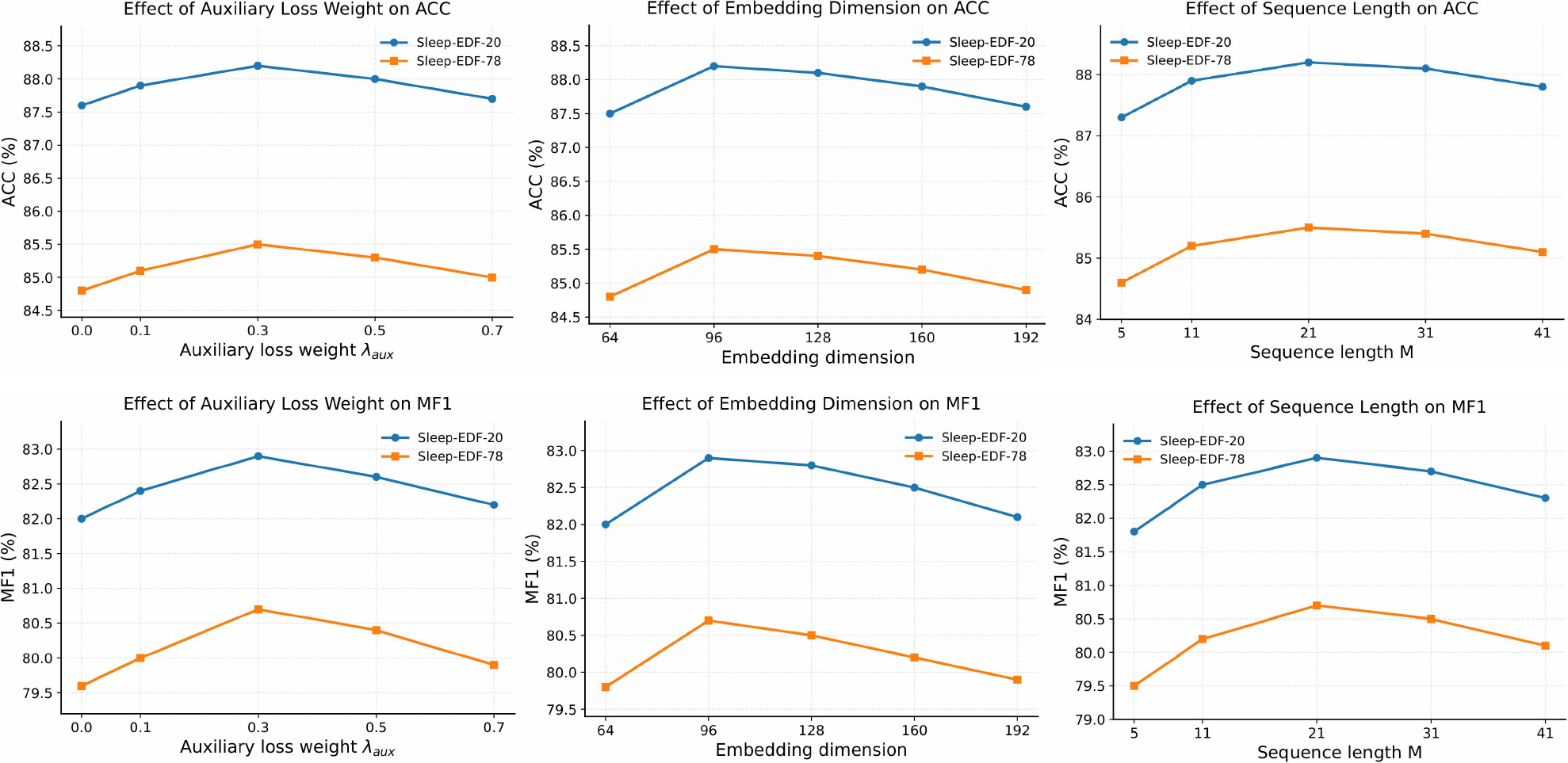}
\caption{Parameter sensitivity analysis of the proposed framework. The figure shows the influence of the contextual sequence length $M$, the embedding dimension, and the auxiliary loss weight $\lambda_{\text{aux}}$ on ACC and Macro-F1 on Sleep-EDF-20 and Sleep-EDF-78.}
\label{fig:param_sensitivity}
\end{figure*}

In addition to class-level and case-level analyses, we also visualize the influence of several important hyperparameters on model performance. As shown in Fig.~\ref{fig:param_sensitivity}, we consider representative parameters such as the contextual sequence length, the Swin window size, and the embedding dimension. The goal of this analysis is not to search exhaustively for the best setting, but rather to examine whether the proposed framework behaves consistently under different parameter choices.

A desirable model should exhibit a relatively stable performance trend within a reasonable parameter range. If the performance changes smoothly rather than drastically across neighboring settings, this indicates that the method is not overly sensitive to hyperparameter tuning. In our case, such behavior would imply that the proposed framework is practically robust and does not rely on a narrow configuration to achieve good results.

At the same time, the curves can also reveal meaningful structural tendencies. For example, if performance first improves and then saturates as the contextual sequence length increases, this suggests that contextual information is beneficial up to a certain range, beyond which redundant temporal context may contribute little additional value. Similarly, changes in the Swin window size can provide insight into the appropriate granularity for local time--frequency modeling. These observations can further support the ablation results in Section~\ref{sec:ablation} and help explain why the final model configuration is reasonable.

\section{Discussion}

The results presented above suggest that the proposed framework provides an effective way to improve automatic sleep staging by strengthening epoch-level representation learning within a hierarchical contextual modeling paradigm. Unlike conventional dual-stream architectures that mainly rely on standard CNN or recurrent encoders for single-epoch representation learning, the proposed method introduces a Swin Transformer into the time--frequency branch to better capture local spectro-temporal structures, cross-region dependencies, and multi-scale patterns. This design yields consistent gains across Sleep-EDF-20, Sleep-EDF-78, and SHHS, indicating that the proposed framework is effective under both medium-scale and large-scale sleep staging settings.

A notable observation is that the proposed method shows clearer advantages on difficult stages, especially N1 and transition-related epochs. This behavior is consistent with the motivation of the model. Since these stages usually contain weaker and more ambiguous discriminative cues, improving the quality of epoch-level representations is particularly important before contextual refinement is applied. The visualization results and ablation studies further support this point, showing that the proposed framework can reduce confusion between neighboring stages and produce more stable sequence-level predictions.

Another important finding is that the hierarchical formulation remains essential even after strengthening the epoch encoder. The results indicate that neither local representation enhancement nor contextual modeling alone is sufficient. Instead, the best performance is achieved when stronger epoch-level features are combined with inter-epoch contextual refinement. This confirms that sleep staging should be treated as a task requiring both local structural evidence and temporal transition modeling.

Despite these encouraging results, several limitations remain. First, the current study is conducted on public benchmark datasets, and further validation on more diverse clinical cohorts would be beneficial. Second, although the proposed method improves N1 recognition, N1 remains the most difficult class overall. Third, the introduction of a Swin-based encoder increases model complexity to some extent, which may need further consideration in lightweight or real-time deployment scenarios.

Overall, the present study shows that enhancing epoch-level representation learning is a practical and effective direction for improving automatic sleep staging. The proposed framework preserves the strengths of hierarchical contextual modeling while providing stronger local structural encoding, which together contribute to more accurate and stable sleep stage prediction.

\section{Conclusion}

Although a conclusion may review the  main points of the paper, do not replicate the abstract as the conclusion. A
conclusion might elaborate on the importance of the work or suggest
applications and extensions.

\section*{Acknowledgment}

\noindent \textbf{Author Contributions:} Conceptualization, C.W. and J.G.; methodology, C.W. and J.G.; software, C.W.; validation, C.W. and J.G.; formal analysis, C.W.; investigation, C.W.; writing---original draft preparation, C.W.; writing---review and editing, J.G.; supervision, J.G. All authors have read and agreed to the published version of the manuscript.

\noindent \textbf{Funding:} This research received no external funding.

\noindent \textbf{Institutional Review Board Statement:} Not applicable.

\noindent \textbf{Informed Consent Statement:} Not applicable.

\noindent \textbf{Data Availability Statement:} The datasets analyzed in this study are publicly available. The Sleep-EDF dataset can be accessed from PhysioNet at https://physionet.org/content/sleep-edfx/. The Sleep Heart Health Study (SHHS) dataset is available from the National Sleep Research Resource (NSRR) at https://sleepdata.org/datasets/shhs. No new data were generated in this study.

\noindent \textbf{Acknowledgments:} In this section you can acknowledge any support given which is not covered by the author contribution or funding sections. This may include administrative and technical support, or donations in kind (e.g., materials used for experiments).

\noindent \textbf{Conflicts of Interest:} The authors declare no conflict of interest.

\begin{IEEEbiography}
[{\includegraphics[width=1in,height=1.25in,clip,keepaspectratio]{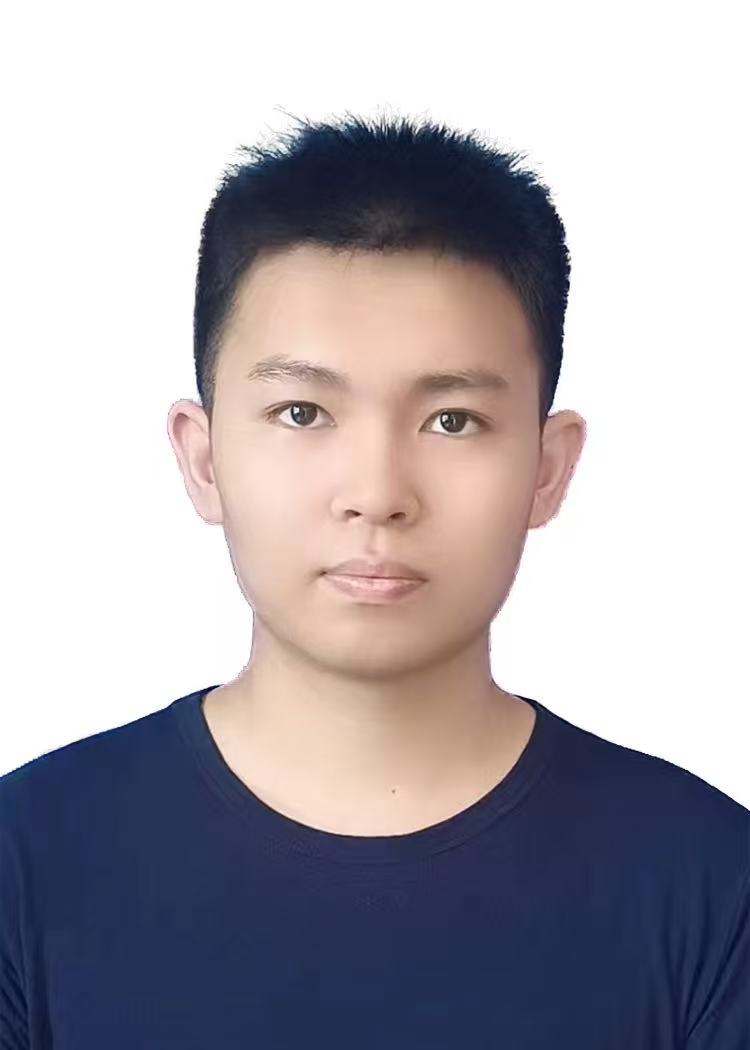}}]{Chongjian Wang}
graduated from Shandong Experimental High School in 2023. He is currently pursuing a bachelor’s degree at the School of Mathematics and Systems Science, Shandong University of Science and Technology. His research interests include point cloud registration in computer vision and electroencephalogram (EEG)–based sleep staging and classification. His publications and academic activities can be found on his Google Scholar profile: \href{https://scholar.google.com/citations?user=LZhu-n0AAAAJ&hl=zh-CN&oi=sra}{Google Scholar}.
\end{IEEEbiography}

\begin{IEEEbiography}
[{\includegraphics[width=1in,height=1.25in,clip,keepaspectratio]{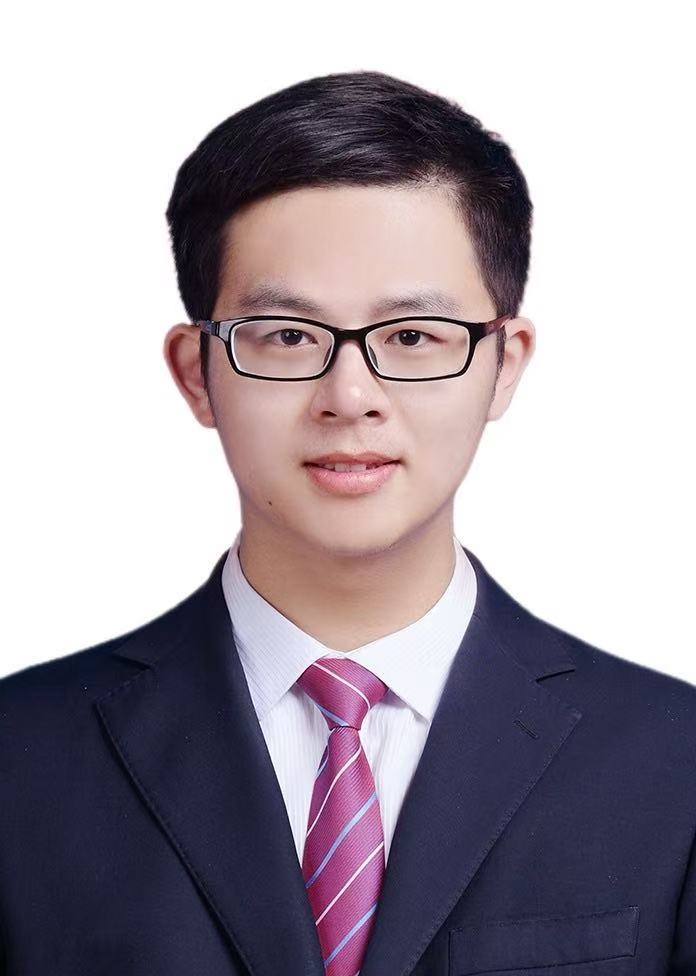}}]{Junjie Gao}
received his Ph.D. degree from the School of Computer Science and Technology, Shandong University, in 2024. He is currently a Lecturer with the School of Artificial Intelligence, Shandong Women’s University, Jinan, China. His research interests include computer vision, computer graphics, and affective analysis. His publications and academic activities can be found on his Google Scholar profile: \href{https://scholar.google.com/citations?user=CTDs13EAAAAJ&hl=zh-CN}{Google Scholar}.
\end{IEEEbiography}

\EOD

\end{document}